\documentclass[preprint,12pt]{elsarticle}

\usepackage{amssymb}
\usepackage{amsmath}

\usepackage{booktabs}
\usepackage[shortlabels]{enumitem}   
\usepackage{amsfonts}
\usepackage{algorithmic}
\usepackage{array}
\usepackage[caption=false,font=normalsize,labelfont=sf,textfont=sf]{subfig}
\usepackage{textcomp}
\usepackage{stfloats}
\usepackage{url}
\usepackage{verbatim}
\usepackage{graphicx}
\usepackage{multirow}
\usepackage{tabularx} 
\usepackage{tabularray}
\usepackage{stfloats}  

\usepackage{microtype}  
\usepackage{url}        
\usepackage{hyperref}   
\usepackage{float}

\AtBeginEnvironment{tabular}{}   
\AtBeginEnvironment{tabular*}{}
\AtBeginEnvironment{array}{}

\SetTblrInner[row]{rowsep=1.4pt}   

\usepackage{caption}        
\journal{ADVANCED ENGINEERING INFORMATICS}

\begin{document}

\begin{frontmatter}



\title{\textbf{HAFO: A Force-Adaptive Control Framework for Humanoid Robots in Intense Interaction Environments}}




\author[1,2]{Chenhui Dong}
\author[3,2]{Haozhe Xu}
\author[3,2]{Wenhao Feng}
\author[1,2,3,4]{Zhipeng Wang}
\author[1,2,3,4]{Yanmin Zhou}
\author[1,2]{Yifei Zhao}
\author[1,2,3,4]{Bin He}

\affiliation[1]{organization={Frontiers Science Center for Intelligent Autonomous Systems, Tongji University},
            city={Shanghai},
            postcode={201109}, 
            country={China}}

\affiliation[2]{organization={National Key Laboratory of Autonomous Intelligent Unmanned Systems},
            city={Shanghai},
            postcode={201109}, 
            country={China}}

\affiliation[3]{organization={Department of Control Science and Engineering, College of Electronics and Information Engineering, Tongji University},
            city={Shanghai},
            postcode={201804}, 
            country={China}}

\affiliation[4]{organization={Shanghai Artificial Intelligence Laboratory},
            city={Shanghai},
            postcode={200030}, 
            country={China}}
\begin{abstract}
Reinforcement learning (RL) controllers have made impressive progress in humanoid locomotion and light-weight object manipulation. However, achieving robust and precise motion control with intense force interaction remains a significant challenge. To address these limitations, this paper proposes HAFO, a dual-agent reinforcement learning framework that concurrently optimizes both a robust locomotion strategy and a precise upper-body manipulation strategy via coupled training. We employ a constrained residual action space to improve dual-agent training stability and sample efficiency. The external tension disturbances are explicitly modeled using a spring-damper system, allowing for fine-grained force control through manipulation of the virtual spring. In this process, the reinforcement learning policy autonomously generates a disturbance-rejection response by utilizing environmental feedback. The experimental results demonstrate that HAFO achieves whole-body control for humanoid robots across diverse force-interaction environments using a single dual-agent policy, delivering outstanding performance under load-bearing and thrust-disturbance conditions, while maintaining stable operation even under rope suspension state. 
\end{abstract}



\begin{keyword}
Humanoid robot \sep Force-adaptive control \sep Whole-body control \sep Spring-damper system \sep Dual-agent reinforcement learning



\end{keyword}

\end{frontmatter}

\section{Introduction}
In recent years, learning-based control solutions have made significant progress in the field of whole-body control for humanoid robots \cite{1,2,3,4,5,6}. By leveraging large-scale parallel simulations and domain randomization, policy networks can effectively learn to handle environmental uncertainties, thus enabling zero-shot transfer to real robots. However, current RL controllers often overlook or oversimplify interactive forces. To address the practical needs of humanoid robots in tasks such as heavy-load manipulation and high-altitude operations, achieving precise and robust motion control is crucial in environments with strong external force disturbances.

When humanoid robots encounter strong collisions or physical contact, reinforcement learning controllers often exhibit instability. Specifically, current RL methods primarily rely on high-level commands, such as movement velocity, root height, or predefined trajectories, which may mitigate some random disturbances. However, when exposed to human intervention or significant environmental contact, the absence of explicit modeling of contact dynamics often hinders the robot's ability to maintain balance, potentially leading to safety hazards \cite{7,8}. For example, in high-altitude operations, ropes are required for support, but their interference can significantly undermine the robot's stability. In heavy-load scenarios, large payloads significantly impair balance, placing considerable demands on the load-bearing capacity.

To enhance the stability of control strategies under human intervention or significant external disturbances, this paper introduces HAFO, a force-adaptive control framework based on explicit dynamic modeling. HAFO employs a dual-agent strategy that decouples the control of the upper and lower bodies. The lower-body strategy focuses on robust locomotion to maintain gait stability under strong external disturbances. The upper-body strategy emphasizes precise and real-time upper-body manipulation under external force disturbances. We utilize a spring-damper mechanical system to model external tension disturbances. Meanwhile, we progressively increase external forces through a curriculum-learning schedule and randomize the point of force application, systematically enhancing the strategy's generalization to various force disturbances. The contributions of this paper are as follows:

\begin{enumerate}[(1)]
	\item A dual-agent reinforcement learning framework is proposed to facilitate the collaborative evolution of the upper and lower bodies in the presence of large external disturbances. The lower-body strategy focuses on ensuring robust locomotion, and the upper-body strategy focuses on precise manipulation. For upper-body training, we introduce a constrained residual action space, which significantly enhances the stability of the two agents during training. Through adversarial training, we achieve stable and efficient coordinated whole body control on the Unitree G1, and successfully extend it to the full-sized humanoid robot Unitree H1-2, further demonstrating its robustness and strong scalability.
	\item A spring-damping dynamic model is developed to explicitly model the external tension force as equivalent spring-damper forces acting on the robot's body. During model training, a progressive curriculum learning strategy is employed, gradually increasing the spring-damper forces to enable the policy to adapt progressively to external disturbances. The reinforcement learning policy autonomously evolves force-adaptive control modes for ground locomotion and aerial suspension, spontaneously generating mode-switching behaviors based on environmental feedback, without relying on explicit state machines or predefined switching logic.
    \item Experimental results demonstrate that HAFO achieves force-adaptive control for humanoid robots under diverse conditions such as heavy load, thrust disturbance and rope suspension using a single dual-agent policy, and exhibits stable and precise motion capabilities under single or dual-hand loads. Moreover, HAFO is the first locomotion-control strategy to achieve stable operation and safe startup under rope-suspension conditions, marking preliminary exploratory work for humanoid robot applications in high-altitude environments.
\end{enumerate}

\section{Related Work}
\subsection{Whole-body control based on reinforcement learning}
Conventional dynamic modeling methods have shown impressive control performance in structured scenarios for legged robots \cite{9,10,11,12,13,14,15,16,17}. However, due to the intrinsic under-actuation of humanoid robots and the variability of real-world environments, it is extremely difficult to establish an accurate dynamics model. Recently, several RL-based solutions have provided new insights for whole-body control of humanoid robots \cite{1,2,3,4,6,7,8,18,19,20,21,22,23,24,25,26,27,28,29,30,31,32}. Common paradigms include lower-RL-upper-IK and integrated-whole-body-RL, which address the challenge of whole-body motion control from different perspectives.

In terms of lower-RL-upper-IK paradigm, Ben et al. \cite{18} proposed the Homie, which employs curriculum learning to dynamically adjust the motion tracking threshold, allowing the policy to progressively adapt to high-difficulty motions during training. By combining large-scale human motion data with reinforcement learning, Cheng et al. \cite{19} encourage the robot's upper body to imitate the reference motion, while relaxing the imitation constraints on the legs, allowing the robot to achieve expressive movements. The Mobile-TeleVision proposed by Lu et al. \cite{23} introduces the predictive motion prior (PMP), which encodes multi-modal motion features through conditional variational autoencoder (CVAE) to guide the reinforcement learning strategy generates dynamic motions. For the integrated-whole-body-RL paradigm, Xie et al. \cite{28} proposed a KungfuBot, which incorporates foot-ground contact states and a motion-phase signal to accurately imitate highly dynamic wushu movements. He et al. \cite{2} introduced the OmniH2O teleoperation system, which integrates a teacher-student distillation framework to transfer knowledge from a privileged teacher policy to the deployed policy, enabling efficient and stable motion control of the robot in real-world environments. He et al. \cite{21} proposed ASAP, which significantly enhances a robot's adaptability in complex environments by leveraging a pretrained motion tracking policy with a residual action model trained using real-world data. Liao et al. \cite{27} designed a scalable and high-quality motion tracking framework called BeyondMimic, which transforms kinematic references into robust, highly dynamic robot motions. Through distillation, it integrates the trained motions into a single policy, enabling the flexible combination of multiple skills during testing and achieving goal-driven control using simple cost functions.

Although whole-body motion control for humanoid robots has made remarkable progress in the research mentioned above, the existing methods still face many challenges. On the one hand, the lower-RL-upper-IK paradigm employs an open-loop control mode for the upper body, which struggles to perceive and adapt to external interference in real time. On the other hand, the integrated-whole-body-RL paradigm controls whole-body joints through a single controller, but due to the weak task goal correlation between upper-body manipulation and lower-body walking, it is prone to overfitting.

\subsection{Force-adaptive control}
Humanoid robots inevitably experience continuous disturbances from external forces in real-world environments. However, existing research on force-adaptive control is relatively scarce. In the field of manipulation, hybrid position and force control methods are commonly employed. Nevertheless, due to the uncertainty of interaction points between humanoid robots and the environment, accurately measuring interaction forces with force sensors becomes challenging, making it difficult to directly apply these methods to legged robots.

Portela \cite{33} et al. first introduced the force-adaptive control objective for quadruped robots, which significantly improved the force interaction capability of the end effector. However, the model requires explicit switching between position mode and force mode, and assumes the system is in a quasi-static state, making it difficult to handle dynamic and unknown disturbances. For humanoid robots, their inherent characteristics, such as a high center of gravity and a small support base, make force-adaptive control particularly complex \cite{34,35,36}. Li et al. \cite{37} introduced a force-adaptive torso-tilt reward, which maintains balance and effectively transmits force through torso tilting and center-of-gravity adjustment, generating greater pulling forces in high-intensity force interaction tasks. Zhang et al. \cite{6} designed a force curriculum, which maximizes the force adaptability while ensuring the feasibility of joint torque, achieving high-intensity force interactions. Kuo et al. \cite{38} proposed a hybrid control method combining the Artificial Rabbit Optimization (ARO) algorithm with an MLP neural network. Through multi-angle impact experiments on flat and sloped terrain, the method can quickly adapt to external forces and effectively maintain balance. Existing model-based methods rely on precise modeling of pre-planned force trajectories, which restricts their generalization capability in unknown environments, rendering them inadequate for adapting to dynamically changing conditions \cite{39}. In contrast, estimation-based approaches \cite{40} are only applicable under quasi-static conditions and fail to effectively address force adaptive control in dynamic loco-manipulation tasks. 

Although some works have studied force adaptive control for humanoid robots, the range of scenarios explored so far remains relatively limited. In this paper, we propose a force-adaptive whole-body control strategy (HAFO), which endows humanoid robots with force-adaptive capabilities in challenging conditions, such as heavy loads,  thrust
disturbance, and tension disturbances using a single
dual-agent policy.

\begin{figure}[t]\centering
	\includegraphics[width=\textwidth]{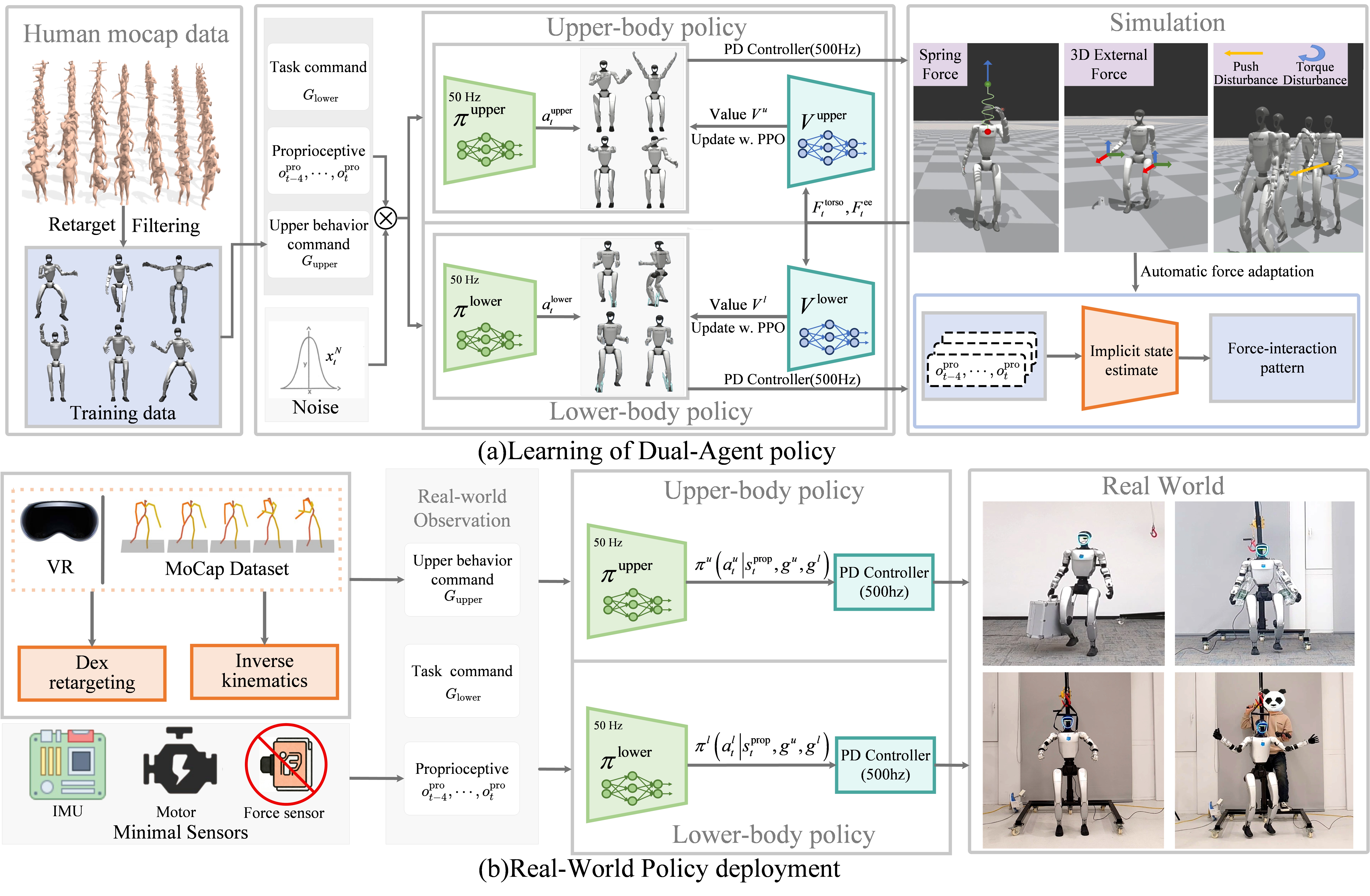}
	\caption{Overview of the HAFO model. (a) Policy Training. A dual-agent strategy with decoupled upper and lower bodies is adopted, where the lower-body policy takes root linear and angular velocities as command inputs, and the upper-body policy uses reference joint trajectories as command inputs. Additionally, various explicit dynamic perturbations are introduced at key locations to enhance the system's robustness and adaptability. (b) Strategy Deployment. A humanoid robot control system based on teleoperation is developed, employing an efficient inverse kinematics algorithm to compute the robot's joint angles in real time with high precision, enabling efficient loco-manipulation tasks. }\label{FIG_1}
\end{figure}

\section{Method}
Humanoid robot motion control requires coordinated movement of whole-body joints. Traditional whole-body control methods rely on a single strategy to output target joint angles for the entire body. However, the lack of strong coupling between upper and lower bodies tasks leads to inefficient sample utilization, making the policy prone to overfitting. To maximize the exploration efficiency of reinforcement learning, this paper proposes HAFO architecture shown in Fig. \ref{FIG_1}, which employs a dual-agent reinforcement learning framework to drive the co-evolution of upper and lower bodies control strategies. During training, various dynamic disturbances are explicitly introduced to ensure stable control.

\subsection{Dual-agent learning framework}

We decompose the total degrees of freedom (DOFs) of the humanoid into upper and lower bodies, such that  $n_{\text {dof }}=n_{\text {dof }}^{\text {upper }}+n_{\text {dof }}^{\text {lower }}$, where $n_{\text {dof }}^{\text {upper }}$ is controlled by upper-body strategy $\pi_{\text {upper}}\left(a_{\text {upper }} \mid \mathrm{s}\right)$, $n_{\text {dof }}^{\text {lower }}$ is controlled by the lower-body strategy $\pi_{\text {lower }}\left(a_{\text {lower }} \mid \mathrm{s}\right)$. 
The dual-agent reinforcement learning strategy is trained using the Proximal Policy Optimization (PPO) algorithm \cite{41} and shares the same state space ${s}_{{t}}=\left[{s}_{{t}}^{\text {prop }}, \mathrm{c}^{l}, \mathrm{~g}^{{u}}\right]$, allowing both agents to consider information from the entire body and achieve coordinated control. 
Among them, the ${s}_{\mathrm{t}}^{\text {prop }}=\left[{q}_{{t}}, \dot{q}_{t}, \omega_{t}, g v_{t}, a_{t-1}\right]$ represents the body perception information of the robot, the $c^{l}=\left[\hat{v}_{x, t}, \hat{v}_{y, t}, \hat{\omega}_{y a w}, \phi_{\text{stance}}\right]$ represents the command information for the lower-body strategy, and the ${g}^{{u}} \in \mathbb{R}^{14}$ represents the command information for the upper-body strategy. The details of the state and action Space are given in Appendix~\hyperref[sec:state-action-space]{A}.

The dual-agent policy is formulated as a dual-objective optimization problem. The value functions $\left(V^{l}, V^{u}\right)$ corresponding to the upper and lower body policies $\left(\pi^{l}, \pi^{u}\right)$, and the rewards obtained can be represented as $\left(r^{l}\left(s, a^{l}, a^{u}\right), r^{u}\left(\mathrm{~s}, \mathrm{a}^{1}, \mathrm{a}^{\mathrm{u}}\right)\right)$. The specific reward functions and RL training parameters can be found in Appendix~\hyperref[sec:Reward Design]{B} and Appendix~\hyperref[sec:RL training]{C}. The dual policies are optimized by maximizing the cumulative rewards \cite{42},

\begin{equation}
\left\{
\begin{array}{l}
\displaystyle \max_{\theta_u} \mathbb{E}_{\pi_{\text{upper}}}\Bigg[\sum_{t=0}^{T-1} \gamma^t r_t\Bigg] \quad \text{(upper policy)} \\[18pt]
\displaystyle \max_{\theta_l} \mathbb{E}_{\pi_{\text{lower}}}\Bigg[\sum_{t=0}^{T-1} \gamma^t r_t\Bigg] \quad \text{(lower policy)}
\end{array}
\right.
\end{equation}

Where $\gamma \in[0,1]$ represents the reward discount factor, $T$ denotes the total number of time steps, $\theta_u$ and $\theta_l$ respectively represent the policy parameters upper and lower bodies.

\subsection{Robust lower-body locomotion under external forces}
Robust locomotion of humanoid robots remains a significant challenge, primarily due to the complexity of the floating base and the instability caused by high inertia. The difficulty is further compounded when payloads are applied to the upper body, increasing the overall inertia and introducing additional modeling uncertainties, which complicate the control design.

For the lower-body strategy $\pi^{l}:C^{l} \times S \mapsto A^{l}$, various external disturbances are introduced during the training process to enhance the robot's robustness, including adversarial upper-body perturbations and high-frequency random pushing disturbances. Additionally, we introduce various domain randomizations during training to improve the model's generalization, with specific details provided in Appendix~\hyperref[sec:Domain randomization]{D}.

\subsection{Upper-body motion generation under external forces}
The demonstration data for the upper-body policy are derived from the AMASS human-motion dataset \cite{43} after feasibility filtering. Following PHC \cite{24}, human poses are retargeted to humanoid poses, and filtering is applied to suppress jitter and discontinuities. Instead of using open-loop replay, we train a reinforcement learning policy for the upper body that adapts to changing environments, avoids self-collisions, and improves whole-body coordination. The upper-body policy $\pi^{\mathrm{u}}: C^{{u}} \times S \mapsto A^{{u}}$ is trained in conjunction with the lower-body policy $\pi^{l}: C^{l} \times S \mapsto A^{l}$, and the state space of the upper-body strategy includes the state of the entire body. To enhance the load capacity of the humanoid robots, disturbances are applied to the robot’s end-effectors in the form of random 3D spatial forces \cite{6}. Similar to the lower-body policy training, various domain randomizations are also incorporated during the upper-body policy training, with specific details provided in Appendix~\hyperref[sec:Domain randomization]{D}.

During the training of the upper-body strategy, samples are randomly selected from the offline dataset $M_{{t}}=\mathrm{q}_{1: T}^{\text {upper }}$ to maintain data variety and prevent overfitting to local patterns. To alleviate the learning difficulty of the lower-body policy, we design a progressive imitation curriculum that guides the network to gradually learn the upper-body reference trajectory, progressing from simple to complex and coarse to fine.
\begin{align}
q_{\text{curr }}^{u}(t) &= q_{0}^{u}(t) + \alpha_{i}\left(q_{\text{tar }}^{u}(t) - q_{0}^{u}(t)\right)
\end{align}
Among them, $q_{0}^{u}(t)$ represents the actual joint angle vector at time $t$, $q_{\text{tar}}^{u}(t)$ is the target joint angle vector of the upper-body at that time $t$, and $\alpha_{i} \in [0,1]$ is the course gain coefficient, which modulates the motion amplitude of the upper-body reference trajectory.

We adopt a constrained residual action space, in which the policy does not directly learn absolute joint targets $\theta_{t}^{\text {target}}$, but instead outputs a corrective offset $\mathrm{a}_{t}^{u}$ relative to the sampled upper-body reference trajectory joint positions $\theta_{t}^{\text {ref}}$. To prevent training collapse caused by adversarial interactions between the dual agents, the output of $\mathrm{a}_{t}^{u}$ is constrained, ensuring that the network’s action outputs remain within a controllable range.
\begin{align}
\theta_{t}^{\mathrm{target}} = \theta_{t}^{\mathrm{ref}} + \mathrm{a}_{t}^{u},\  \text{s.t. } |\mathrm{a}_{t}^{u}| \leq \Delta
\end{align}
\subsection{Virtual spring-damper system}
In complex and unpredictable environments, a precise and stable control system is crucial for achieving task objectives. To improve the robustness of humanoid robots in the presence of unknown external disturbances, this paper introduces a tension-force modeling mechanism based on spring-damper system \cite{44,45,46}, explicitly incorporating the tension force interaction process into the training loop, aiming to regulate the robot's body dynamics through a virtual spring-damper model, the corresponding dynamic model is given by
\begin{align}
m\ddot{x}=K_{\mathrm{p}}\left(x_{\text {des }}-x\right)+K_{\mathrm{d}}\left(\dot{x}_{\text {des }}-\dot{x}\right)+f_{\text {ext }}
\end{align}

Where $K_{p}$ and $K_{d}$ represent the stiffness and damping of the virtual spring-damper system, $\left(x_{\mathrm{des}}, \dot{x}_{\mathrm{des}}\right)$ denotes the desired position and velocity, and $f_{\text {ext }}$ represents the external force.

\begin{figure}[t]\centering
	\includegraphics[width=1.0\textwidth]{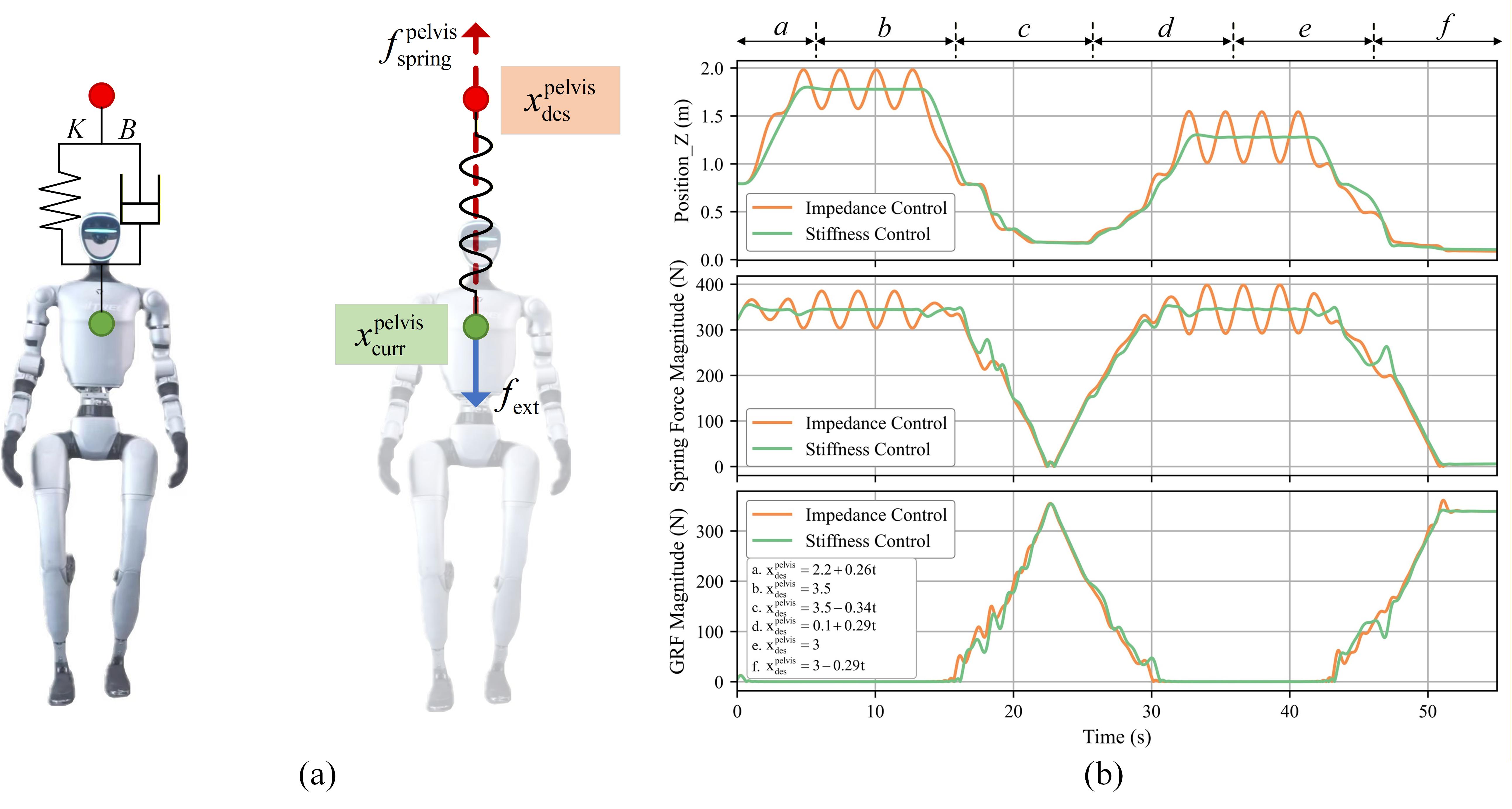}
	\caption{Spring-damper model and performance analysis. (a) Spring-damper model schematic on the humanoid robot. (b)Pelvis position, spring force, and ground-reaction force versus a specified $x_{\mathrm{des}}^{\text {pelvis }}$ displacement for Spring-damper and stiffness models.}\label{FIG_2}
\end{figure}

Leveraging a spring-damper model, we accurately capture the mechanics of external tensile force. By progressively adjusting the $x_{\mathrm{des}}^{\text {pelvis }}$, the tensile force is tuned to achieve stable position control, as shown in Fig. \ref{FIG_2}. Comparative tests show that stiffness control produces pronounced oscillations relative to spring-damper model, confirming the critical role of the damping term in enhancing stability and precision. We integrate the spring-damper system into the training loop, applying the force application position, cycle, and dropout timing randomization to effectively enhance the policy's generalization capability.

\begin{figure}[t!]\centering
	\includegraphics[width=1.0\textwidth]{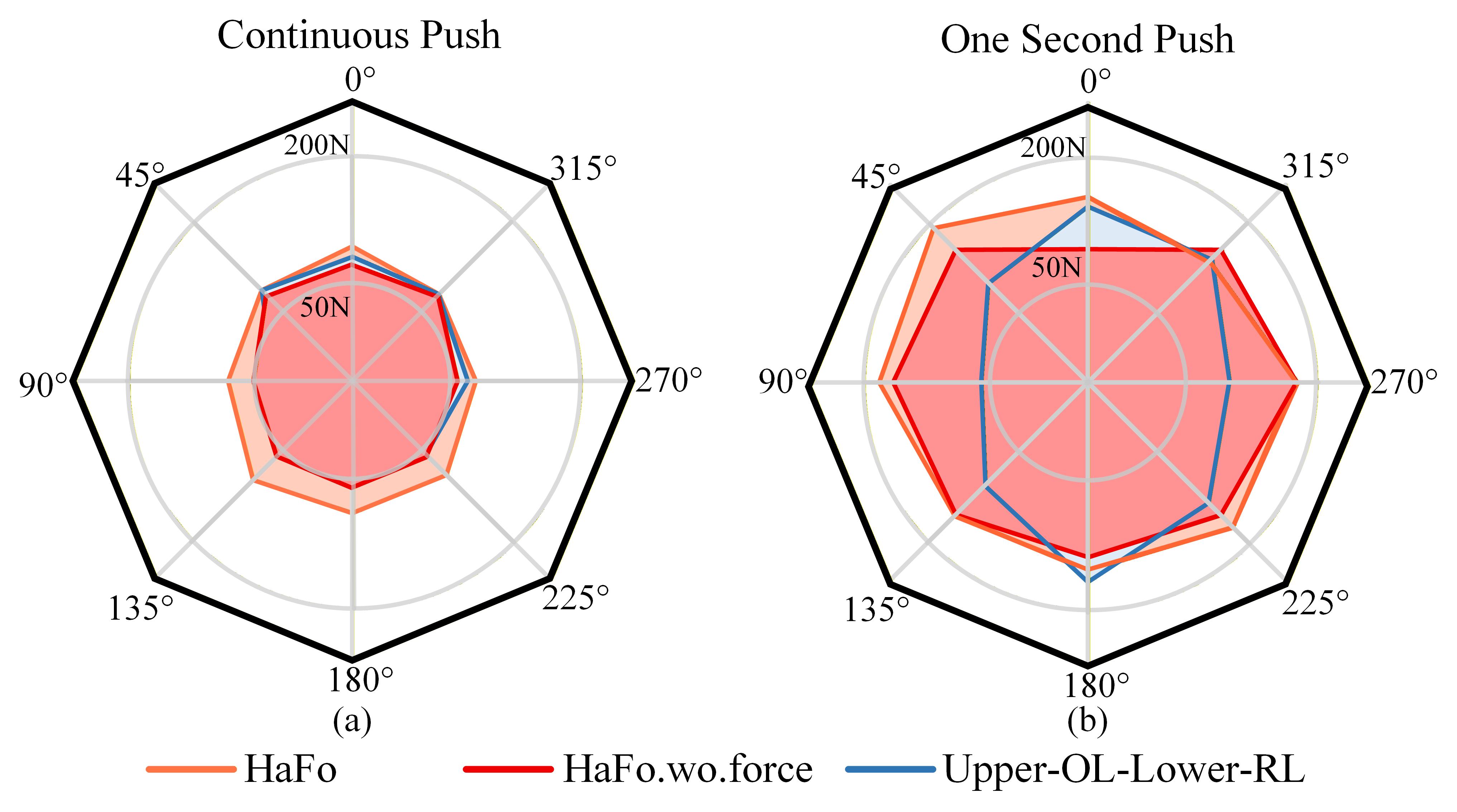}
	\caption{Unitree G1 Humanoid robot sim2sim results. We evaluate the model's performance in diverse intense-interaction environments.  (a) Achieving stable control in a suspension scenario. (b) Achieving a smooth, autonomous transition from a suspended state to a ground movement state. (c) Swinging both arms while each hand bears a 30 N external force. (d) Punching while each hand bears a 30 N external force. (e) Rapidly recovering balance after experiencing a lateral 250 N thrust disturbance lasting 0.2 seconds. (f) Rapidly recovering balance after experiencing a longitudinal 250 N thrust disturbance lasting 0.2 seconds. }\label{FIG_3}
\end{figure}

\section{Simulation Experiments}

In this section, we conduct a series of force-interaction simulation experiments and comprehensively evaluated the effectiveness of the proposed HAFO framework through quantitative comparisons with the baseline methods. All simulations are conducted in MuJoCo \cite{47} via zero-shot transfer from Isaac Gym \cite{48}. The cross-platform validation approach effectively mitigates overfitting risks inherent to a single simulator, thereby effectively verifying policy robustness. As shown in Fig. \ref{FIG_3}, we display the evaluation results of HAFO across various force disturbance scenarios, providing comprehensive validation of the policy's force adaptability.

\subsection{Force adaptive control under hand-induced disturbances}

To evaluate the stability performance of the model under external disturbances, we apply forces of different directions and magnitudes to the robot's end-effectors in the simulation environment, with the force magnitude incrementally increasing from 10N to 50N. During the force disturbance process, the robot executes random velocity commands (linear velocity range: [-0.5, 0.5] m/s, angular velocity range: [-0.5, 0.5] rad/s) while performing diverse upper-limb motion sequences sampled from a subset of the AMASS dataset \cite{43}. We test key indicators such as the upper-body motion tracking error and velocity tracking error. Motion tracking error is quantified by calculating the difference between the actual joint angles and the target joint angles, and velocity tracking error is quantified by calculating the difference between the actual velocity and the target velocity. 

Upper-body motion tracking error:
\begin{align}
E_{\text{error}}^{\text{upper}} &= \frac{1}{T} \sum_{t=1}^{T} \left| q_{t}^{\text{upper*}} - q_{t}^{\text{upper}} \right|
\end{align}

Velocity tracking error:
\begin{align}
E_{\text{error}}^{\text{vel}} &= \frac{1}{T} \sum_{t=1}^{T} \left| v_{t}^{\text{lin,ang*}} - v_{t}^{\text{lin,ang}} \right|
\end{align}

The velocity tracking error $\mathrm{E}_{\text {error }}^{\text {vel }}(\mathrm{m} / \mathrm{s})$ and upper-body motion tracking error $\mathrm{E}_{\text {error }}^{\text {upper }}(\mathrm{rad})$ are computed as the average values over a complete iteration. To evaluate the effectiveness of the proposed force-adaptive control strategy, two representative benchmarks are selected:

(i)	upper-OL-lower-RL: a decoupled policy with open-loop upper-body control and RL-driven lower-body control \cite{18}.

(ii) upper-FIX-lower-RL: a lower-body motion strategy with the upper body fixed \cite{49}.

\begin{table}[t]
\caption{Policy Tracking Performance Evaluation. We evaluated the upper-body motion tracking performance and lower-body velocity tracking performance of different policies under three external force magnitudes applied to each hand (N-Force: 10 N, S-Force: 30 N, M-Force: 50 N). These metrics reflect the force adaptability of the policies under external disturbance, where "/" indicates task failure.}
\label{table 1}
\setlength{\tabcolsep}{3.5pt}   
\centering
\fontsize{8.5pt}{10pt}\selectfont
\begin{tabular}{lcccccc}
\toprule[1pt]
\multirow{2}{*}[-0.8ex]{Methods}
& \multicolumn{3}{c}{$\mathrm{E}_{\text{tracking}}^{\text{upper}}\downarrow$}
& \multicolumn{3}{c}{$\mathrm{E}_{\text{tracking}}^{\text{root}}\downarrow$} \\
\cmidrule(lr){2-4}\cmidrule(lr){5-7}
& S-Force & N-Force & L-Force & S-Force & N-Force & L-Force \\
\midrule
upper-OL-lower-RL  & 0.36±0.03 & 0.74±0.04 & 1.23±0.07 & \textbf{0.32±0.04} & 0.53±0.10 & 1.52±0.22 \\
upper-FIX-lower-RL & 0.55±0.06 & /         & /         & 0.52±0.09 & /         & /         \\
HAFO w.o. DA       & 0.34±0.08 & 0.66±0.11 & 1.32±0.17 & 0.50±0.02 & 0.55±0.03 & 0.77±0.07 \\
HAFO w.o. Force    & 0.42±0.03 & 0.60±0.03 & 1.36±0.11 & 0.36±0.03 & 0.54±0.08 & 0.92±0.06 \\
HAFO(Ours)               & \textbf{0.22±0.04} & \textbf{0.41±0.03} & \textbf{0.46±0.04} & 0.33±0.02 & \textbf{0.48±0.03} & \textbf{0.53±0.05} \\
\bottomrule[1pt]
\end{tabular}
\end{table}

Furthermore, ablation experiments are conducted to quantify the contribution of each HAFO module. We sequentially ablate (i) the dual-agent architecture, retaining only the lower-body agent (w.o. DA), and (ii) the force-adaptative module (w.o. Force). As shown in Table \ref{table 1}, comparative experiments demonstrate that HAFO achieves superior tracking performance, both in terms of velocity tracking and upper-body angle following. The upper-FIX-lower-RL policy simplifies training by fixing the upper-body and relying exclusively on lower-body control for locomotion. However, it fails to guarantee system stability when the upper-body executes dynamic motions. Notably, HAFO demonstrates a more pronounced performance advantage as the hand-applied external force increases. Furthermore, ablation results demonstrate that removing either module leads to performance degradation. The random upper-body motion introduced during training serves as an adversarial perturbation source, potentially compelling the lower-body policy to learn more generalized dynamic compensation mechanisms. Meanwhile, force-interaction curriculum learning effectively enhances the dual-agent's robustness and adaptability under perturbations.

\subsection{Force adaptive control under rope suspension}


\begin{table}[t]
  \centering
  \fontsize{9pt}{10pt}\selectfont
  \setlength{\tabcolsep}{25pt}   
  \caption{Comparison of different methods in terms of action smoothness and upper-body motion tracking error.}
  \label{table 2}
  \begin{tabular}{lcc}
    \toprule[1pt]
    Methods                    & $\Delta a\downarrow$ & $E_{\text{tracking}}^{\text{upper}}\downarrow$ \\
    \midrule
    upper-OL-lower-RL          & 2.49$\pm$2.38        & 0.27$\pm$0.11 \\
    upper-FIX-lower-RL         & 5.66$\pm$4.50        & 0.29$\pm$0.15 \\
    HAFO w.o.\ DA              & 1.60$\pm$0.49        & 0.27$\pm$0.24 \\
    HAFO-w/o-force             & 4.52$\pm$1.77        & 1.16$\pm$0.66 \\
    HAFO (Ours)                & \textbf{0.38$\pm$0.16} & \textbf{0.20$\pm$0.04} \\
    \bottomrule[1pt]
  \end{tabular}
\end{table}

We conduct a comprehensive evaluation of the strategy’s force adaptability under suspension, focusing on upper-body motion tracking error, action smoothness. Upper-body motion tracking error $\mathrm{E}_{\text {tracking }}^{\text {upper }}\downarrow \text { (rad) }$ is quantified by the deviation between target joint angles and actual joint angles, while action smoothness $\Delta \mathrm{a}\downarrow(\mathrm{~m} / \mathrm{s})$ is measured by motion smoothness. Under dynamic disturbances during suspension, significant variations occur in the robot's motion states, including angular velocity, linear velocity, body posture, and foot-ground contact. HAFO maintains excellent performance across various lifting and landing scenarios through real-time environmental feedback and adaptive behavior adjustment. As shown in Table \ref{table 2}, when using HAFO policy, the upper-body motion tracking error is 0.38, and action smoothness is 0.20, significantly enhancing force adaptability in rope suspension scenarios and outperforming existing strategies. Notably, HAFO exhibits substantial compliance and stability during lifting and landing transitions, fully validates that the dual-agent collaborative mechanism combined with force curriculum learning enables systematic and precise identification of suspension dynamics characteristics. Furthermore, ablation studies demonstrate that, despite undergoing force curriculum learning, HAFO w.o. DA employs open-loop control for the upper limbs, lacking active adaptability and leading to significant performance degradation.

\begin{figure}[t]
\centering
\includegraphics[width=0.8\columnwidth]{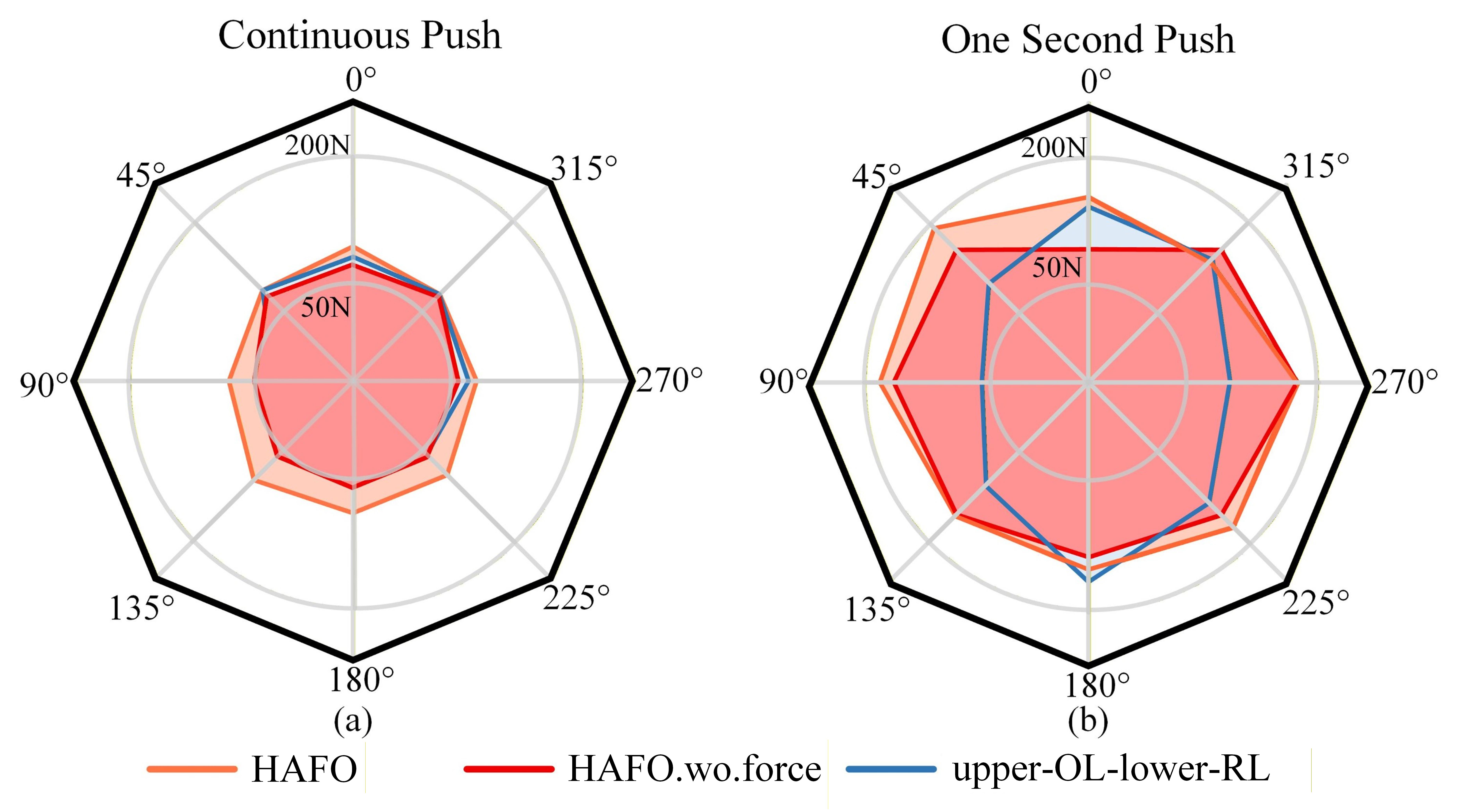}
\caption{Policy performance under external thrust disturbances. (a) Performance under continuous constant-force disturbances applied from multiple directions. (b) Performance under one-second transient force disturbances from multiple directions.}
\label{FIG_4}
\end{figure}

\subsection{Force-Adaptive Control under Thrust Disturbance}

To systematically assess policy robustness against thrust perturbations, we conduct two disturbance rejection experiments: applying sustained constant forces and one-second transient forces along eight directions in the horizontal plane, progressively increasing force magnitude until loss of stability. As shown in Fig. \ref{FIG_4}, HAFO demonstrates greater tolerance to external disturbances in most directions compared to HAFO w.o. force, and the baseline strategy, indicating that adversarial upper-body training and the force curriculum both improve the robot's force adaptation under thrust perturbations.

\section{Real-world experiments}
\begin{figure}[H]\centering
	\includegraphics[width=\textwidth]{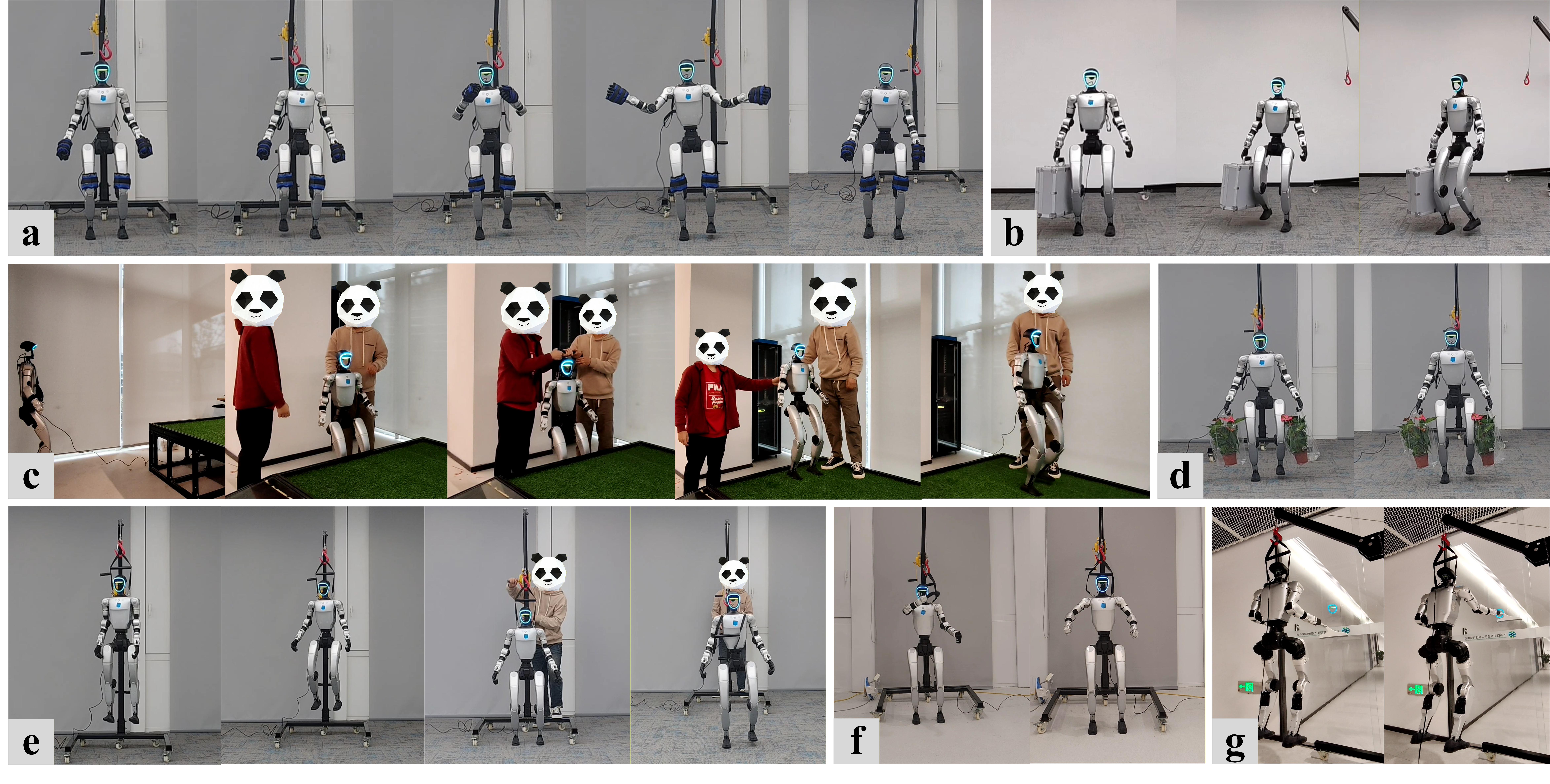}
	\caption{HAFO enables force-adaptive control for humanoid robots across multiple scenarios. (a) Achieving whole-body control with sandbag loads. (b) Walking while carrying a box with one hand. (c) Being lifted onto an elevated platform and resuming locomotion. (d) Walking while carrying flower pots with both hands. (e) Deploying the policy in a suspended state. (f) Performing dance movements. (g) Cleaning windows while being suspended by ropes.}
    \label{FIG_5}
\end{figure}

We deployed HAFO strategy on physical Unitree G1 humanoid robot and conduct a series of hardware experiments. As shown in Fig. \ref{FIG_5}, the HAFO controller completes multiple force interaction tasks, collectively validating the technical performance and practical applicability of our method on the physical robot.

\subsection{Loco-manipulation under hand loads}
To evaluate the robot’s motion performance under load, we conduct experiments with both single-hand and dual-hand loads. In the experiment, we record the velocity tracking error and upper-body motion tracking error respectively. As shown in Table \ref{table 3}, whether the robot carries the load with one hand or both, HAFO maintains the lowest velocity-tracking and upper-limb joint errors, outperforming the baseline methods.

\begin{table}
\caption{Policy Tracking Performance Evaluation on the Real-world robot. We evaluate upper-body motion tracking error and lower-body velocity tracking error of different policies across difficulty hand-loading modes (1 kg load in one hand, 1 kg load in each hand).}
\label{table 3}
\setlength{\tabcolsep}{10pt}   
\centering
\fontsize{9pt}{10pt}\selectfont
\begin{tabular}{lcccc}
\toprule[1pt]
\multirow{2}{*}[-0.8ex]{Methods}   
& \multicolumn{2}{c}{One hand loads}
& \multicolumn{2}{c}{Two hand loads} \\
\cmidrule(lr){2-3}\cmidrule(lr){4-5}
& $\mathrm{E}_{\text{tracking}}^{\text{upper}}\downarrow$
& $\mathrm{E}_{\text{tracking}}^{\text{root}}\downarrow$
& $\mathrm{E}_{\text{tracking}}^{\text{upper}}\downarrow$
& $\mathrm{E}_{\text{tracking}}^{\text{root}}\downarrow$ \\
\midrule
upper-OL-lower-RL & 0.28±0.05 & 0.49±0.03 & 0.53±0.08 & 0.50±0.04 \\
HAFO w.o. DA      & 0.45±0.08 & 0.58±0.05 & 0.64±0.12 & 0.57±0.05 \\
HAFO w.o. Force   & 0.40±0.06 & 0.43±0.08 & 0.54±0.07 & 0.44±0.11 \\
HAFO(Ours)              & \textbf{0.16±0.05} & \textbf{0.40±0.04} & \textbf{0.39±0.03} & \textbf{0.41±0.03} \\
\bottomrule[1pt]
\end{tabular}
\end{table}

\subsection{Stable control under suspension}
\begin{figure}
    \centering
    \includegraphics[width=0.6\columnwidth]{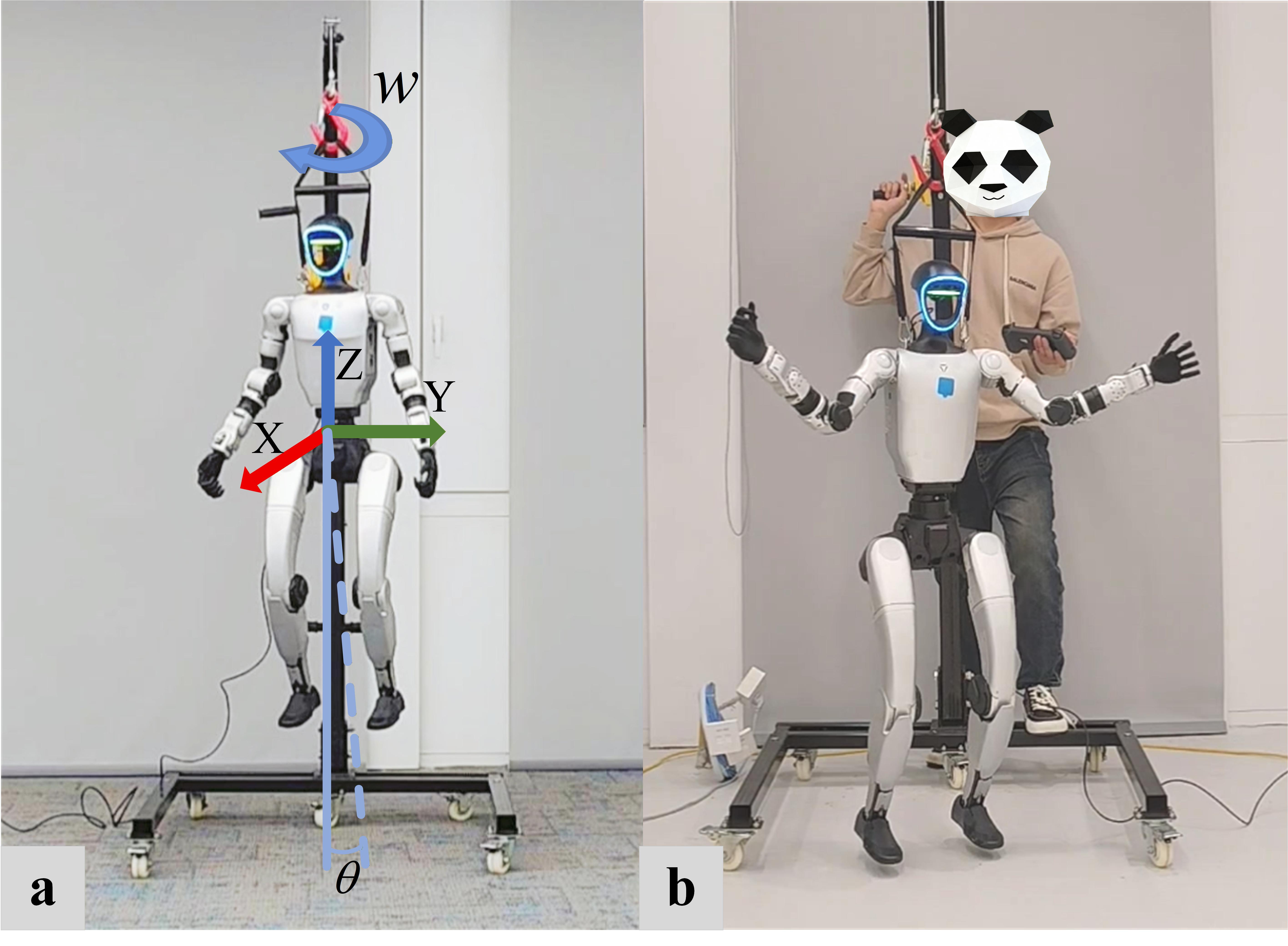}
    \caption{Motion analysis of the robot under rope suspension conditions. (a) Under the influence of rope swing, the robot experiences dynamic changes in linear velocity $v_{\text {lin}}$, angular velocity $\omega_{\text {yaw}}$, and body tilt angle $\theta$, and other motion parameters during the suspension process. (b) The HAFO strategy integrates environmental feedback, enabling the velocity-commanded motion policy to achieve adaptive control in the suspended state.}
    \label{FIG_6}
\end{figure}

Traditional policy initialization schemes typically require robots to sequentially execute the series of procedures: foot-ground contact, posture adjustment, policy execution, and rope release. This process is relatively cumbersome, and sustained external disturbances from the rope readily leads to robot instability. This experiment validates the feasibility of suspended startup. The HAFO controller can initiate directly from a rope-suspended state and rapidly reach steady state, significantly simplifying the policy deployment process. Humanoid robots are typically controlled based on high-level signals such as target velocity and foot contact commands. Under the dynamic interference of rope tension, significant changes occur in the robot's locomotion states, including linear velocity, angular velocity, body posture, and foot-ground contact, as shown in Fig. \ref{FIG_6}. The baseline strategy misinterprets the normal swinging of the rope as robot instability, triggering emergency actions that cause violent oscillations. In contrast, the HAFO policy leverages environmental feedback for adaptive compensation, robustly maintaining locomotion stability.

\begin{table}[t]
  \centering
  \fontsize{9pt}{10pt}\selectfont
  \caption{Policy performance evaluation across different task scenarios. We assess motion smoothness and upper-body tracking error for each task scenario.}
  \label{table 4}
  \setlength{\tabcolsep}{23pt}          
  \begin{tabular}{lcc}
    \toprule[1.0pt]
    Tasks            & $\Delta a\downarrow$ & $\mathrm{E}_{\text{tracking}}^{\text{upper}}\downarrow$ \\
    \midrule
    Hang on start      & 0.19±0.01           & 0.13±0.01 \\
    High-altitude work & 0.43±0.05           & 0.21±0.03 \\
    \bottomrule[1.0pt]
  \end{tabular}
\end{table}

To verify the force adaptability in the rope suspension scenario, we simulate typical suspended operation conditions and systematically test the whole process of lifting, operating and landing. During the lifting phase, the robot is lifted from the ground, and its posture stability is monitored in real time. After entering the operation phase, the robot performs the teleoperation tasks in the suspended state. Finally, the robot achieves a stable bipedal stance following touchdown. We compare with the existing scheme, and finally only the HAFO completed the final experiment. The results are shown in Table \ref{table 4}, we show the application potential of HAFO in high-rise outdoor dangerous scenes, and provide new ideas and methods for the practical application of robots in the future. 


\section{Conclusion}
In this paper, we propose HAFO, a force-adaptive control framework for humanoid robots. HAFO employs a dual-agent control strategy and explicitly introduces various dynamic disturbances, achieving stable motion control and precise motion imitation. During the model training, a spring-damper system is established, and fine-grained tension control of the tension force is achieved through manipulation of virtual springs. Additionally, HAFO employs asymmetric Actor-Critic architecture wherein the critic-network is granted full access to privileged state variables, thereby furnishing a more accurate approximation of state-action value. Experimental results demonstrate that the proposed strategy enables the humanoid robots to exhibit strong adaptability under challenging conditions such as heavy load, thrust disturbance and rope suspension using a single dual-agent policy, providing robust support for solving substantial force interaction tasks in real-world applications.




\bibliographystyle{elsarticle-num-names}
\bibliography{BIB}           

\appendix
\setcounter{table}{0}
\setcounter{figure}{0}

\newpage
\section{State and Action Space}
\label{sec:state-action-space}
This section provides a detailed description of the state and action spaces for the dual-agent policy. The state space for both the upper and lower body strategies is identical, consisting of the whole-body state set. The dimensions of the action space for the upper and lower body strategies correspond to the number of degrees of freedom (DOF) in the humanoid robot's upper and lower bodies, respectively. Specific details are shown in Table \ref{table A.1}.

\begin{table}[H]
  \centering
  \fontsize{9pt}{10pt}\selectfont
  \caption{State and Action space description.}
  \label{table A.1}
  \setlength{\tabcolsep}{10pt}         
  \begin{tabular}{lccc}
    \toprule[1.0pt]
    Term              & Lower dim & Upper dim & Whole dim \\
    \midrule
    Base ang vel      & 3         & 3         & 3         \\
    Projected gravity & 3         & 3         & 3         \\
    Commands          & 4 (Vel cmd + Stand cmd) & 14 (Motion angles) & 18 \\
    Dof pos           & 29        & 29        & 29        \\
    Dof vel           & 29        & 29        & 29        \\
    Actions           & 15 (Lower) & 14 (Upper) & 29 (Whole) \\
    \midrule
    Total dim         & 83        & 92        & 111       \\
    \bottomrule[1.0pt]
  \end{tabular}
\end{table}

We adopt an asymmetric Actor-Critic architecture for reinforcement learning training. In this architecture, the Critic network has access to privileged information, including the base linear velocity and external force values, which allows it to more accurately evaluate the value of the current state and provide more precise feedback signals to the Actor network.

\section{Reward Design}
\label{sec:Reward Design}
This paper adopts a reward function design similar to that in \cite{2,6,19}, while making targeted additions and modifications to certain reward components to better accommodate the needs of full-body control tasks and force adaptation control. These adjustments aim to enhance both the training effectiveness and the robustness of the strategy. The specific reward components are detailed in Table \ref{table B.2}.

\begin{table}[H]
\centering
\caption{Definition of the additional reward function}
\label{table B.2}
\fontsize{8.5pt}{11pt}\selectfont
\newcommand{\Is}{\mathbb{I}(\|F_{\text{spring}}\|)}  
\begin{tblr}{
  colspec = {l X[1,l] c},          
  cells   = {l},                   
  cell{2}{1} = {c=3}{c},           
  cell{16}{1}={c=3}{c},            
  hline{1-3,16-17,22} = {-}{},
  hline{1} = {-}{wd=1.1pt},    
  hline{22} = {-}{wd=1.1pt},    
}
Term & Expression & Weight \\
Regularization Reward && \\
Orientation
  & $\|\mathbf{g}_{xy}^{\text{root}}\|_2^2$
  & -4.0 \\
Torso Orientation
  & $\|\mathbf{g}_{xy}^{\text{torso}}\|_2^2$
  & -7.0 \\
Lower-Body Action Rate
  & $\|a_t^{\text{lower}}-a_{t-1}^{\text{lower}}\|^2$
  & -0.2 \\
Feet Orientation
  & $\|\mathbf{g}_{xy}^{\text{feet}}\|_2^2$
  & -4.0 \\
Termination
  & $\mathbb{I}_{\text{termination}}$
  & -350 \\
Feet Parallel
  & $\operatorname{Var}(D)$
  & -2.0 \\
Lower-Body Stand Still
  & $\|q_t^{\text{upper}}-q_t^{\text{default}}\|\times\Is>0$
  & -2.2 \\
Base Height
  & $\|\mathbf{h}-\mathbf{h}_{\text{target}}\|_2^2\times\Is=0$
  & -10 \\
Additional Torques
  & $\|\boldsymbol{\tau}\|_2^2\times\Is>0$
  & -0.0001 \\
Additional Dof Vel
  & $\|\dot{\mathbf{q}}\|_2^2\times\Is>0$
  & -0.008 \\
Additional Dof Acc
  & $\|\ddot{\mathbf{q}}\|_2^2\times\Is>0$
  & -0.000011 \\
Additional Action Rate
  & $\|a_t-a_{t-1}\|_2^2\times\Is>0$
  & -0.01 \\
Horizontal Angular Velocity
  & $\|\boldsymbol{\omega}_{xy}\|_2^2\times\Is=0$
  & -1.0 \\
Task Reward && \\
Linear X Velocity
  & $\exp[-\|v_t^{x*}-v_t^{x}\|^2/0.25]\times\Is=0$
  & 2.0 \\
Linear Y Velocity
  & $\exp[-\|v_t^{y*}-v_t^{y}\|^2/0.25]\times\Is=0$
  & 2.0 \\
Angular Yaw Velocity
  & $\exp[-\|v_t^{\text{ang}*}-v_t^{\text{ang}}\|^2/0.25]\times\Is=0$
  & 6.0 \\
Upper-Body Dofs Position
  & $\exp[-\|q_t^{\text{upper}*}-q_t^{\text{upper}}\|^2/0.25]$
  & 4.0 \\
Lower-Body Stand Still
  & $\exp[-\|q_t^{\text{lower}}-q_t^{\text{default}}\|^2/0.25]\times\Is>0$
  & 3.0 \\
\end{tblr}
\end{table}

\section{RL training}
\label{sec:RL training}
We use Proximal Policy Optimization (PPO) \cite{41} as the base algorithm, and Table \ref{table C.3} presents the detailed training hyperparameters.

\begin{table}[H]
  \centering
  \fontsize{9pt}{10pt}\selectfont
  \caption{RL controller hyperparameters.}
  \label{table C.3}
  \setlength{\tabcolsep}{35pt}   
  \begin{tabular}{lcc}
    \toprule[1.0pt]
    Hyperparameter       & \multicolumn{2}{c}{Value} \\
    \midrule
    Clip param           & \multicolumn{2}{c}{0.2} \\
    Discount factor ($\gamma$) & \multicolumn{2}{c}{0.99} \\
    Value loss coef      & \multicolumn{2}{c}{1.0} \\
    Entropy coef         & \multicolumn{2}{c}{0.01} \\
    Actor learning rate  & \multicolumn{2}{c}{0.001} \\
    Critic learning rate & \multicolumn{2}{c}{0.001} \\
    Weight decay         & \multicolumn{2}{c}{0.01} \\
    Max grad norm        & \multicolumn{2}{c}{1.0} \\
    Desired KL           & \multicolumn{2}{c}{0.01} \\
    Num steps per env    & \multicolumn{2}{c}{24} \\
    MLP size             & \multicolumn{2}{c}{[512, 256, 128]} \\
    \bottomrule[1.0pt]
  \end{tabular}
\end{table}

\section{Domain randomization}
\label{sec:Domain randomization}
To enhance the robustness of the policy and reduce the performance gap between the simulation and real environments, we applied domain randomization to key parameters in the simulation. This approach allows the policy better adapt to unknown factors when transferred to the real environment. The specific domain randomization parameters and their ranges are listed in Table \ref{table D.4}.

\begin{table}[H]
\caption{Domain randomization parameters and their ranges.}
\label{table D.4}
\fontsize{9pt}{10pt}\selectfont
\centering
\begin{tblr}{
  hline{1-2,8} = {-}{},
  hline{1} = {-}{wd=1.0pt},    
  hline{8} = {-}{wd=1.0pt},    
  colsep = 18pt,        
}
Term            & Value                                                        \\
Ground friction & $\mathcal{U}(0.25,1.25)$                                   \\
Kp and Kd       & $\mathcal{U}(0.9,1.1)$                                     \\
Base mass       & $\mathcal{U}(-1,3)\,\mathrm{kg}$                           \\
Link mass       & $\mathcal{U}(0.9,1.2)\times\text{default}\,\mathrm{kg}$    \\
Base CoM        & $\mathcal{U}(-0.01,0.01)\,\mathrm{cm}$                     \\
Control delay   & $\mathcal{U}(0,20)\,\mathrm{ms}$                           \\
\end{tblr}
\end{table}

\section{Extend HAFO to Large-size Humanoid Robots}
\label{sec:Extend HAFO to Large-size Humanoid Robots}
\begin{figure}[htbp]\centering
	\includegraphics[width=1.0\textwidth]{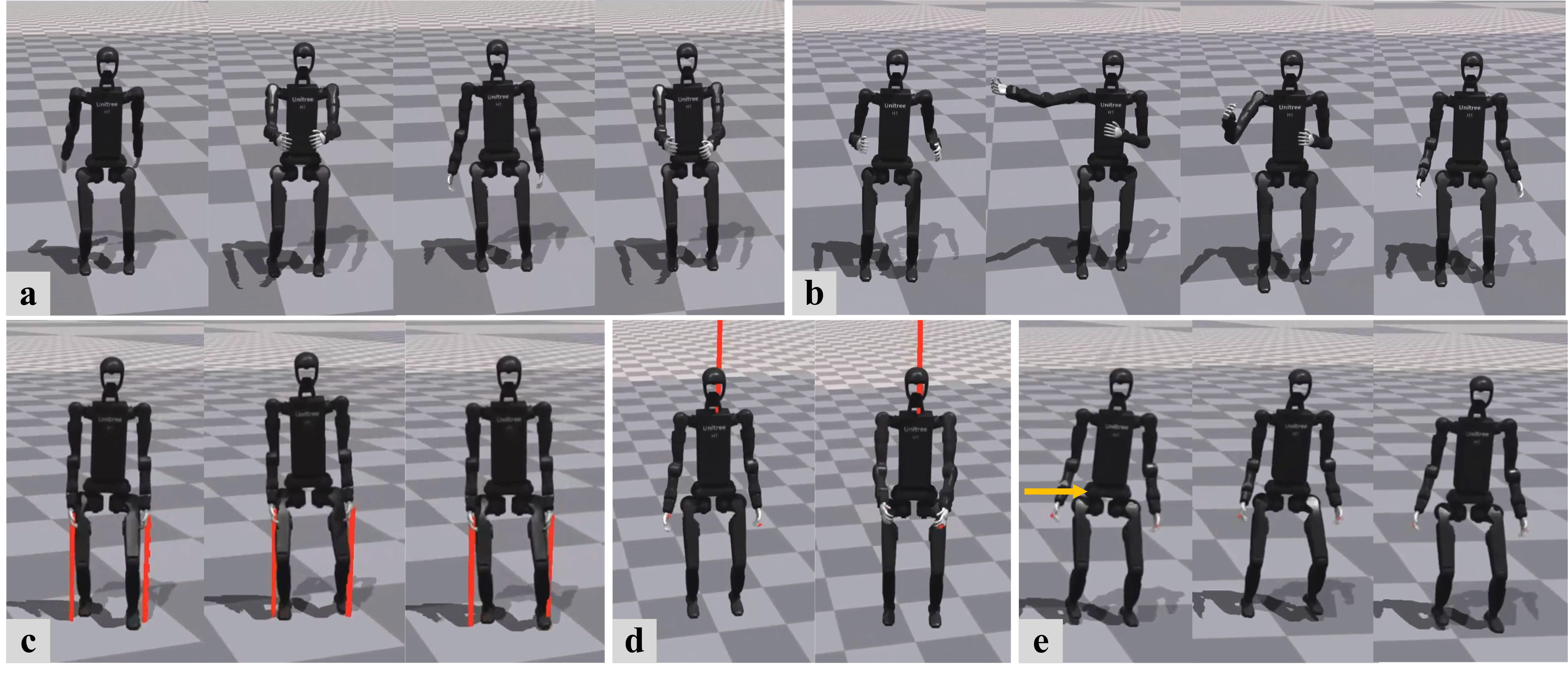}
	\caption{Multi-scenario validation of the HAFO policy on the unitree H1-2 humanoid robot. (a) Swinging the robot's upper body. (b) Balancing under Punching motion. (c) Walking with a 100 N external force applied to each hand. (d) Achieving stable control in the suspended state. (e) Rapidly recovering balance after being subjected to a thrust disturbance.}\label{AppendixFig}
\end{figure}

As shown in Fig. \ref{AppendixFig}, we further extend the HAFO strategy to the full-scale humanoid robot H1-2 (height: 1.78 m, weight: 70 kg). The increased mass and limited actuators make controlling the robot more challenging. However, with minimal parameter adjustments, stable control is achieved within the simulation environment, thereby validating the strategy's strong generalization and scalability.

\end{document}